\documentclass[preprint]{sig-alternate-05-2015}
\usepackage{blindtext, graphicx}
\usepackage{pgfplots}
\pgfplotsset{compat=1.3}
\usetikzlibrary{intersections}
\usepackage{url}
\usepackage{amsmath}
\usepackage{tabularx}
\usepackage{booktabs}
\usepackage{algorithm}
\usepackage{algpseudocode}
\usepackage{subcaption} 
\usepackage[font=small,skip=2pt]{caption}
\usepackage{subfig}
\usepackage{siunitx}
\usepackage{textcomp}
\usepackage{microtype}
\usepackage{glossaries}
\newacronym{FPGA}{FPGA}{Field Programmable Grid Array}
\newacronym{CNN}{CNN}{Convolutional Neural Network}
\newacronym{BNN}{BNN}{Binarized Neural Network}
\newacronym{OFM}{OFM}{Output Feature Map}
\newacronym{IFM}{IFM}{Input Feature Map}
\newacronym{OCM}{OCM}{On-Chip Memory}
\newacronym{MVU}{MVTU}{Matrix--Vector--Threshold Unit}
\newacronym{SWU}{SWU}{Sliding Window Unit}
\newacronym{TU}{TU}{Thresholding Unit}
\newacronym{PU}{PU}{Pooling Unit}
\newacronym{II}{II}{initiation interval}
\newacronym{PE}{PE}{Processing Element}
\newacronym{NN}{NN}{Neural Network}
\newacronym{FPS}{FPS}{frames per second}
\newacronym{HLS}{HLS}{High-Level Synthesis}
\newacronym{ILSVRC}{ILSVRC}{ImageNet Large Scale Visual Recognition Competition}
\newacronym{TOPS}{TOPS}{teraoperations per second}
\newacronym{GOP}{GOPS}{billion operations}
\newacronym{GFLOP}{GFLOP}{billon floating point operations}
\newcommand{\dblquotes}[1]{``#1''}
\usepackage{multirow}

\usepackage{todonotes}

\tikzset{
	declare function={
		mysign(\x) = (and(\x<=0, 1) * -1) +
		       (and(\x>0, 1) * 1);
		}
	}

\newcommand{\PowerWall}{$P_{\mathrm{wall}}$}
\newcommand{\PowerFPGA}{$P_{\mathrm{chip}}$}

\newcommand{\OurScheme}{\textsc{Finn}}
\newcommand{\OurSchemeFull}{FINN: A Framework for Fast, Scalable Binarized Neural Network Inference}
\newcommand{\approxtilde}[1]{{\raise.17ex\hbox{$\scriptstyle\sim$}}#1}

\toappear{\emph{To appear in the 25th International Symposium on Field-Programmable Gate Arrays, February 2017.}}

\begin{document}
\CopyrightYear{2017} 
\setcopyright{acmlicensed}
\conferenceinfo{FPGA '17,}{February 22 - 24, 2017, Monterey, CA, USA}
\isbn{978-1-4503-4354-1/17/02}\acmPrice{\$15.00}
\doi{http://dx.doi.org/10.1145/3020078.3021744}

%
\title{\OurSchemeFull{}}

%
%
%
%

\numberofauthors{1} 
\author{
Yaman Umuroglu\textsuperscript{*\dag}, Nicholas J. Fraser\textsuperscript{*\ddag}, Giulio Gambardella\textsuperscript{*}, Michaela Blott\textsuperscript{*},\\
Philip Leong\textsuperscript{\ddag}, Magnus Jahre\textsuperscript{\dag} and Kees Vissers\textsuperscript{*}\\
       \affaddr{\textsuperscript{*}Xilinx Research Labs};
       \affaddr{\textsuperscript{\dag}Norwegian University of Science and Technology};
       \affaddr{\textsuperscript{\ddag}University of Sydney}\\
       \email{yamanu@idi.ntnu.no}
}


\maketitle

\begin{abstract}
Research has shown that convolutional neural networks contain significant redundancy, and high classification accuracy can be obtained even when weights and activations are reduced from floating point to binary values. In this paper, we present \OurScheme{}, a framework for building fast and flexible FPGA accelerators using a flexible heterogeneous streaming architecture. By utilizing a novel set of optimizations that enable efficient mapping of binarized neural networks to hardware, we implement fully connected, convolutional and pooling layers, with per-layer compute resources being tailored to user-provided throughput requirements. On a ZC706 embedded FPGA platform drawing less than 25 W total system power, we demonstrate up to 12.3 million image classifications per second with 0.31~\textmu s latency on the MNIST dataset with 95.8\% accuracy, and 21906 image classifications per second with 283 \textmu s latency on the CIFAR-10 and SVHN datasets with respectively 80.1\% and 94.9\% accuracy.
To the best of our knowledge, ours are the fastest classification rates reported to date on these benchmarks.



\end{abstract}
\section{Introduction}
\glspl{CNN} have dramatically improved in recent years, their performance now exceeding that of other visual recognition algorithms~\cite{krizhevsky2012imagenet}, and even surpassing human accuracy on certain problems \cite{deeplearningoverview, planet}.
They are likely to play an important role in enabling ubiquitous machine vision and intelligence on all kinds of devices, but a significant computational challenge remains. 
Modern \glspl{CNN} may contain millions of floating-point parameters and require billions of floating-point operations to recognize a single image.
Furthermore, these requirements tend to increase as researchers explore deeper networks.
For instance, AlexNet~\cite{krizhevsky2012imagenet} (the winning entry for \gls{ILSVRC} \cite{ilsvrc} in 2012) required 244 MB of parameters and 1.4 \gls{GFLOP} per image, while VGG-16~\cite{simonyan2014very} from \gls{ILSVRC} 2014 required 552MB of parameters and 30.8 \gls{GFLOP} per image.

%
While the vast majority of \glspl{CNN} implementations use floating point parameters, a growing body of research demonstrates this approach incorporates significant redundancy. 
Recently,  it has been shown \cite{binarynet, ResiliencyUnderQuantization, xnornet, bitwiseneuralnet, dorefa} that neural networks can classify accurately using one- or two-bit quantization for weights and activations.
%
Such a combination of low-precision arithmetic and small memory footprint presents a unique opportunity for fast and energy-efficient image classification using \glspl{FPGA}.
\glspl{FPGA} have \emph{much} higher theoretical peak performance for binary operations compared to floating point, while the small memory footprint \emph{removes} the off-chip memory bottleneck by keeping parameters on-chip, even for large networks.
\glspl{BNN}, proposed by Courbariaux~et~al.~\cite{binarynet}, are particularly appealing since they can be implemented almost entirely with binary operations, with the potential to attain performance in the \gls{TOPS} range on \glspl{FPGA}.

In this work, we propose \OurScheme{}, a framework for building scalable and fast \gls{BNN} inference accelerators on \glspl{FPGA}.
\OurScheme{}-generated accelerators can perform millions of classifications per second with sub-microsecond latency, thereby making them ideal for supporting real-time embedded applications such as augmented reality, autonomous driving and robotics. 
Compute resources can be scaled to meet a given classification rate requirement.
We demonstrate \OurScheme{}'s capabilities with a series of prototypes for classifying the MNIST, SVHN and CIFAR-10 benchmark datasets. 
Our classification rate results surpass the best previously published results by over $48\times$ for MNIST, $2.2\times$ for CIFAR-10 and $8\times$ for SVHN.
To the best of our knowledge, this is the fastest reported neural network inference implementation on these datasets.
The novel contributions are:
\begin{itemize}
    \item Quantification of peak performance for \glspl{BNN} on \\ \glspl{FPGA} using a roofline model.
    \item A set of novel optimizations for mapping \glspl{BNN} onto \gls{FPGA} more efficiently.
    \item A \gls{BNN} architecture and accelerator construction tool, permitting customization of throughput.
    \item A range of prototypes that demonstrate the potential of \glspl{BNN} on an off-the-shelf \glspl{FPGA} platform.
\end{itemize}
The rest of this paper is organized as follows: Section 2 provides background on \glspl{CNN}, \glspl{BNN}, and their hardware implementations.
Section 3 discusses \glspl{BNN} accuracy and peak performance on \glspl{FPGA}.
Section 4 describes \OurScheme{}'s architecture and optimizations. Section 5 presents the experimental evaluation, and Section 6 concludes the paper.
\section{Background}


\subsection{Convolutional Neural Networks}

This work is focused on \textit{supervised} learning,
in which the goal is to find a function, $g(\mathbf{x}_i)$, which approximates a mapping  $\mathbf{x}_i \to y_i\ \forall\ i$, 
where $\{\mathbf{x}_i, y_i\}$ is an input/output pair known as a training example.
A \textit{multilayer perceptron} is a type of artificial neural network which
has its neurons arranged in multiple layers, with neurons taking the output of all neurons of the previous layer as inputs.
Mathematically, the output, $a_{l,n}$, for the $n^{th}$ neuron in the $l^{th}$ layer of a fully connected network is calculated as follows:
\begin{equation}
a_{l,n} = f_{act}(\sum_{s=0}^{S_l}w_{l,n,s}a_{l-1,s} + b_{l,n})\ \ \mathrm{,}
\label{eq:fully_connected_inference}
\end{equation}
where $w_{l,n,s}$ is weight of the $s^{th}$ synapse connected to the input of the $n^{th}$ neuron in the $l^{th}$ layer, $b_{l,n}$ is a bias term, $f_{act}$ is the activation function, and $S_l$ is the number of synapses connected to each neuron in the $l^{th}$ layer.
Popular activation functions include:
the hyperbolic tangent function, $f_{act}(a) = tanh(a)$; and
the rectified linear unit (ReLU), $f_{act}(a) = max(0,a)$. Furthermore, only the {\emph{inference} problem is studied, the parameters, $w$, being assumed to have been learned offline.


Convolutional neural networks~\cite{mnist} (CNNs) are a variant of multilayer perceptrons, in which a layer only receives inputs from  a small \emph{receptive field} of the previous layer. This approach greatly reduces the number of parameters involved and allows local features (e.g., edges, corners) to be found~\cite{mnist}.
A basic 2D convolutional layer in a neural network is similar to a fully connected layer except that:
a) each neuron receives an image as inputs and produces an image as its output (instead of a scalar);
b) each synapse learns a small array of weights which is the size of the convolutional window; and
c) each pixel in the output image is created by the sum of the convolutions between all synapse weights and the corresponding images.
%
%

The output of the $l^{th}$ convolutional layer, which takes as input $S_l$ images of dimension $R_l \times C_l$, the pixel, $p_{l,n,r,c}$, at location $(r, c)$ of the $n^{th}$ output image is calculated as follows:
\begin{equation}
p_{l,n,r,c} = f_{act}(\sum_{s=0}^{S_l}\sum_{j=0}^{J_l}\sum_{k=0}^{K_l}w_{l,n,s,j,k}p_{l-1,n,r+j,c+k})\ \ \mathrm{,}
\label{eq:convolution_calc}
\end{equation}
where $J_l \times K_l$ are the dimensions of the convolution window.
As discussed in Section~\ref{sec:architecture}, a 2D convolutional layer can be reduced to a matrix multiply followed by an elementwise activation function.
CNN topologies are composed from a few common primitives: convolutional layers, \textit{pooling} layers and fully connected layers.


\label{sec:BkgPooling}

Pooling layers can be considered as simple downsamplers of 2D images.
A basic max pooling layer divides an image into small sub-tiles of a given window size and then replaces each sub-tile with its largest element. An average pooling
layer is similar but uses the average function instead of max.



\subsection{Binary Neural Networks}
Although floating point numbers are a natural choice for handling the small updates that occur during neural network training, the resulting parameters can contain a lot of redundant information \cite{han2015deep}. 
%
One of several possible dimensions possessing redundancy is precision \cite{ResiliencyUnderQuantization}.
An extreme case are \glspl{BNN} in which
some or all the arithmetic involved in computing the outputs are constrained to single-bit values.
We consider three aspects of binarization for neural network layers: binary input activations, binary synapse weights and binary output activations.
If all three components are binary, we refer to this as \emph{full binarization}, and the cases with one or two components as \emph{partial binarization}.

Kim~and~Smaragdis~\cite{bitwiseneuralnet} consider full binarization with a predetermined portion of the synapses having zero weight, and all other synapses with a weight of one.
They report 98.7\% accuracy with fully-connected networks on the MNIST dataset, and observe that only XNOR and bitcount operations are necessary for computing with such neural networks.
XNOR-Net by Rastegari~et~al.~\cite{xnornet} applies convolutional \glspl{BNN} on the ImageNet dataset with topologies inspired by AlexNet, ResNet and GoogLeNet, reporting top-1 accuracies of up to 51.2\% for full binarization and 65.5\% for partial binarization.
DoReFa-Net by Zhou~et~al.~\cite{dorefa} explores reduced precision during the forward pass as well as the backward pass, and note that this opens interesting possibilities for training neural networks on FPGAs.
Their results includes configurations with partial and full binarization on the SVHN and ImageNet datasets,
including best-case ImageNet top-1 accuracies of 43\% for full and 53\%  for partial binarization.

Finally, the work by Courbariaux~et~al.~\cite{binarynet} describes how to train fully-connected and convolutional networks with full binarization and batch normalization layers, reporting competitive accuracy on the MNIST, SVHN and CIFAR-10 datasets. Training for this work was performed using their open source implementation.
We use the acronym \gls{CNN} to refer to conventional or non-binarized neural networks for brevity throughout the rest of this paper.

\subsection{Neural Networks in Hardware}
A great deal of prior work on mapping neural networks to hardware exist both for FPGAs and as ASICs.
We refer the reader to the work by Misra and Saha~\cite{surveyannhw} for a comprehensive survey.
We cover a recent and representative set of works here, roughly dividing them into four categories  based on their basic architecture:
1) a single processing engine~\cite{ovtcharov2015accelerating,zhang2015optimizing,chen2016eyeriss,YodaNN}, usually in the form of a systolic array, which processes each layer sequentially;
2) a streaming architecture~\cite{venieris2016fpgaconvnet,alemdar2016ternary}, consisting of one processing engine per network layer;
3) a vector processor~\cite{farabet2009cnp} with instructions specific to accelerating the primitives operations of convolutions; and
4) a neurosynaptic processor~\cite{esser2016convolutional}, which implements many digital neurons and their interconnecting weights.

\emph{Systolic arrays:} Zhang~et~al.~\cite{zhang2015optimizing} describes a single processing engine style architecture, using theoretical roofline models tool to design accelerators optimized for the execution of each layer.
Ovtcharov~et~al.~\cite{ovtcharov2015accelerating} implement a similar style architecture, but achieved a 3$\times$ speedup over Zhang~et~al.~\cite{zhang2015optimizing}.
Eyeriss by Chen~et~al.~\cite{chen2016eyeriss} use 16-bit fixed point rather than floating point, and combine several different data reuse strategies.
Each 2D convolution is mapped to 1D convolutions across multiple processing engines, allowing for completely regular access patterns for each processing element.
The authors report that their data reuse provides 2.5$\times$ better energy efficiency over other methods.
YodaNN by Andri~et~al.~\cite{YodaNN} have a similar design as Zhang~et~al.~\cite{zhang2015optimizing} but explore binary weights for fixed sized windows.

\emph{Streaming architectures:} Venieris and Bouganis~\cite{venieris2016fpgaconvnet} proposed a synchronous dataflow (SDF) model for mapping CNNs to FPGAs, which is a similar approach to ours. The main difference is that our design is optimized for BNNs while their design targets conventional CNNs.
Their designs achieve up to 1.62$\times$ the performance density of hand tuned designs.
Alemdar~et~al.~\cite{alemdar2016ternary} implement fully-connected ternary-weight neural networks with streaming and report up to 255K frames per second on the MNIST dataset, but concentrate on the training aspect for those networks.



\emph{Vector processors:} Farabet~et~al.~\cite{farabet2009cnp} describe a programmable ConvNet Processor (CNP), which is a RISC vector processor with specific macro-instructions for CNNs including 2D convolutions, 2D spatial pooling, dot product and an elementwise non-linear mapping function.
The authors also created a tool to compile a high level network description into host code which is used to call the CNP. 



\emph{Neurosynaptic processors:} TrueNorth~\cite{esser2016convolutional} is a low power, parallel ASIC with 4096 neurosynaptic cores,
each implementing 256 binary inputs, 256 neurons and a 256 $\times$ 256 array of synapses. An internal spiking router can connect any input on any core to any neuron on any core, allowing many network topologies to be implemented on fixed hardware.

The authors are not aware of any publication that investigates mapping of fully binarized neural networks onto FPGAs.
In comparison to prior art, the binary network inference engine can significantly increase classification rates, while reducing power consumption and minimizing latency.
This currently comes at the cost of a small drop in accuracy for larger networks,
however we believe
a) there are use cases that do not require the highest level of accuracy, or can be solved with smaller networks (such as classification of playing cards or handwritten digits~\cite{mnist}) and
b) that the accuracy can be improved by increasing network sizes~\cite{ResiliencyUnderQuantization}. This last point is a ongoing topic in machine learning research.


\section{\gls{BNN} Performance and Accuracy}
\subsection{Estimating Performance Using Rooflines}
\label{subsec:back-rooflines}

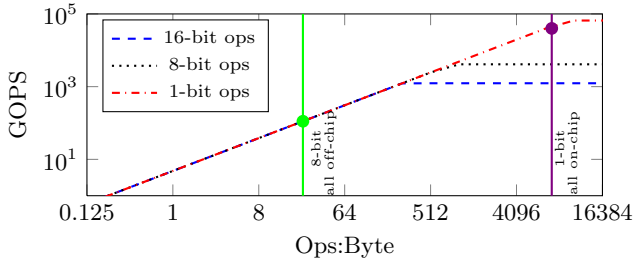
\begin{figure}
\begin{tikzpicture}
\begin{loglogaxis}[
	width=\columnwidth,
	height=4cm,
    xlabel={Ops:Byte},
    ylabel={GOPS},
    log basis x=2,
    xticklabels={0.125, 1, 8, 64, 512, 4096, 16384, 32768},
    xmin=0.125, xmax=32768,
    ymin=1, ymax=100000,
    legend pos={north west},
    legend style={
    	font=\scriptsize
    },
    legend columns=1
]

%
%
\addplot[
    dashed, thick, 
    color=blue,
    mark=none,
    ]
    table [x=OperationalIntensity, y=ZU19EG-HP, col sep=comma] {csv-graph/Roofline-updated.csv};
    \label{16-bit precision-ZU19EG};
\addplot[
    dotted, thick,
    color=black,
    mark=none,
    ]
    table [x=OperationalIntensity, y=ZU19EG-QP, col sep=comma] {csv-graph/Roofline-updated.csv};
    \label{8-bit precision-ZU19EG};
\addplot[
    dashdotted, thick,
    name path global=1bit,
    color=red,
    mark=none,
    ]
    table [x=OperationalIntensity, y=ZU19EG-1b, col sep=comma] {csv-graph/Roofline-updated.csv};
    \label{1-bit precision-ZU19EG}; 
\addplot [color=violet,thick, name path global=DoreFanet] coordinates {
		(9602,1)
		(9602,100000)};
    \label{AlexNet};
\addplot [color=violet, thick, mark=*] coordinates {
		(9602,40000)};   

\addplot [color=green,thick, name path global=AlexNet8BitOffChip] coordinates {
		(23.28,1)
		(23.28,100000)};
    \label{AlexNet8BitOffChip};
\addplot [color=green, thick, mark=*] coordinates {
		(23.28,111.74)};

\draw (5, 3) node[rotate=90]{\tiny 8-bit};
\draw (5.5, 2.7) node[rotate=90]{\tiny all off-chip};

\draw (13.5, 3) node[rotate=90]{\tiny 1-bit};
\draw (14.0, 2.7) node[rotate=90]{\tiny all on-chip};
		
\legend{16-bit ops, 8-bit ops, 1-bit ops}
\end{loglogaxis}

\end{tikzpicture}
\caption{Roofline model for a ZU19EG.}
\label{fig:roofline_ku115}
\end{figure}

To estimate and compare \gls{BNN} performance with fixed-point \gls{CNN}, we use a \emph{roofline model}~\cite{roofline} which considers memory bandwidth, peak computational performance and arithmetic intensity (the number of mathematical operations performed for each byte of off-chip memory read or written).
The intersection of the roofline curve with a vertical line for a particular arithmetic intensity, gives the theoretical peak performance point, which is either \emph{compute-bound} or \emph{memory-bound}.
In particular, we consider the binarized \cite{dorefa, xnornet} and 8-bit fixed-point \cite{fpganclnclconvnet} implementations of the popular AlexNet \cite{krizhevsky2012imagenet}, both of which require 1.4 \gls{GOP} to classify one image.

Using the methodology described in \cite{fpgaroofline}, we develop a roofline model for a Xilinx Zynq UltraScale+ ZU19EG \gls{FPGA}\footnote{We assume 4.8 GB/s off-chip memory bandwidth, 350~MHz clock and the following operation cost function: 2.5 LUTs for 1-bit, 40 LUTs for 8-bit, 8 LUTs and 0.5 DSPs for 16-bit.}.
The resulting roofline model is depicted in Figure~\ref{fig:roofline_ku115}. 
We first observe that the \gls{FPGA}'s compute-bound performance is 66 \gls{TOPS} for binary operations, which is about $16\times$ higher compared to 8-bit and $53\times$ higher compared to 16-bit fixed point operations. 
However, reaching the compute-bound peak is only possible if the application is not memory-bound.
The compact model size of \glspl{BNN} provides another key benefit. 
Since the binarized AlexNet requires only 7.4~MB of parameters (compared with 50 MB for 8-bits), the entire neural network model can be kept in on-chip memory.
The arithmetic intensities for the binarized and 8-bit fixed point AlexNet variants are shown with vertical lines. 
Thus, the \gls{BNN} is almost able to reach the computational peak, while the peak performance of the fixed-point \gls{CNN} is bound by the memory bandwidth.
Based on these observations, with a design that reaches 75\% of the peak, we estimate a throughput of  $0.75 \cdot \frac{66~\mathrm{\gls{TOPS}}}{1.4~\mathrm{\gls{GOP}}} \approx 35000$  images per second.

Using the same model, it should be possible to extend the comparison to CPUs and GPUs, but little data is available on peak binary synaptic operation performance since \glspl{BNN} are relatively new.
For instance, \cite{binarynet} mentions 6 cycles per 32 synapses (64 binary operations) on recent NVIDIA GPUs, which would yield a computational peak of about 26 \gls{TOPS} on a Tesla K40 with 2880 cores running at 875~MHz, and 16666 images per second for binarized AlexNet.



\subsection{Accuracy--Computation Tradeoffs}
\label{subsec:accuracy}
A tradeoff between network size, precision and accuracy exists~\cite{ResiliencyUnderQuantization} so
if one would like to achieve a certain classification accuracy for a particular problem, which approach leads to the most efficient solution?
1) A regular ANN with floating point precision?
2) A larger network, but a BNN?
To gain more insight into this issue, we conducted a set of experiments on the MNIST dataset that compare accuracy of floating point and binary precision for the same topology.
The binary networks are obtained via replacing regular layers by their binary equivalents, as described by Courbariaux~et~al.~\cite{binarynet}.
We also binarize the input images for the \gls{BNN} as our experiments show that input binarization works well for MNIST.
Since the space of possible network topologies that can be trained is infinite, we adopted the approach in \cite{ResiliencyUnderQuantization} to simplify the problem.
We fix the network topology to a 3 hidden layer, fully connected network while scaling the number of neurons in each layer, and plot the resulting accuracy in Table~\ref{tab:accuracy} along with the number of parameters and operations per frame.
A few trends are apparent for this problem and network configuration space:
1) similar to what was found in by Sung~et~al.~\cite{ResiliencyUnderQuantization}, as the network size increases, the difference in accuracy between low precision networks and floating point networks decreases; and
2) in order to achieve the same level of accuracy as floating point networks, BNNs require 2--11$\times$ more parameters and operations.
\begin{table}
\centering
\caption{Accuracy results - BNN vs NN.}
\label{tab:accuracy}
\scriptsize
\resizebox{\linewidth}{!}{
\begin{tabular}{l|cccc}
\toprule
 & Binary   & Float & &\\
Neurons/layer & Err. (\%) & Err. (\%) & \# Params & Ops/frame\\
\midrule
128 & 6.58 & 2.70 & 134,794 & 268,800\\
256 & 4.17 & 1.78 & 335,114 & 668,672\\
512 & 2.31 & 1.25 & 932,362 & 1,861,632\\
1024 & 1.60 & 1.13 & 2,913,290 & 5,820,416\\
2048 & 1.32 & 0.97 & 10,020,874 & 20,029,440\\
4096 & 1.17 & 0.91 & 36,818,954 & 73,613,312\\
\bottomrule
\end{tabular}
}
\end{table}
Note that we show the accuracy for networks trained using 32-bit floating point numbers, but it is likely that this could be reduced to 8-bit fixed point without a significant change in accuracy~\cite{iandola2016squeezenet}.
Our \gls{BNN} performance estimates from Section \ref{subsec:back-rooflines} suggest a $16\times$ speedup for \gls{BNN} over 8-bit fixed point, which is greater than the 2--11$\times$ increase in parameter and operation size.
Thus, we expect that \glspl{BNN} with comparable accuracy will be faster than fixed-point networks, even though they may require more parameters and operations.

\section{\glspl{BNN} on Reconfigurable Logic}
\label{sec:architecture}

\subsection{Architecture}

\begin{figure}[t]
		\centering
		\includegraphics[width=0.8\linewidth]{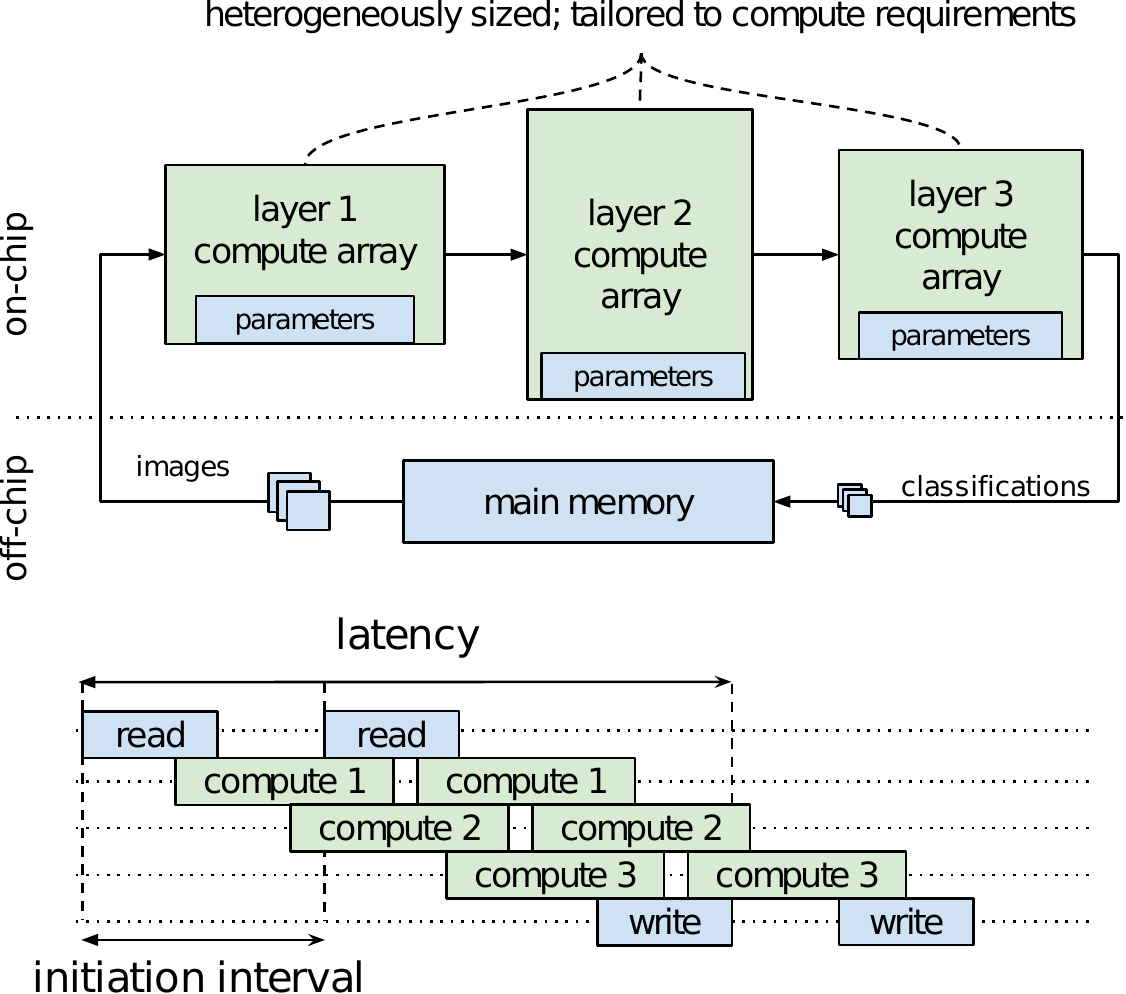}
		\caption{Heterogeneous streaming.}
		\label{fig:ArchStreaming}
\end{figure}

We adopted a \emph{heterogeneous streaming} architecture as shown in Figure \ref{fig:ArchStreaming}, for this work.
We build a custom architecture for a given topology rather than scheduling a operations on top of a fixed architecture.
Separate compute engines are dedicated to each layer, which communicate via on-chip data streams.
Each engine starts to compute as soon as the previous engine starts to produce output.
Additionally, owing to the compact model size of \glspl{BNN}, all neural network parameters are kept in on-chip memory.
This avoids most accesses to off-chip memory, minimizes the \emph{latency} (the time to finish classifying one image) by overlapping computation and communication, and minimizes the \emph{initiation interval}: a new image can enter the accelerator as soon as the first compute array is finished with the previous image.
The separate mapping of layers to compute arrays also enables heterogeneity.
By tailoring compute arrays separately for each layer's requirements, we can avoid the \dblquotes{one-size-fits-all} inefficiencies and reap more of the benefits of reconfigurable computing.
This requires a different bitfile when the neural network topology is changed but we consider this an acceptable cost for the performance gains obtained.

A \gls{BNN} accelerator may have various constraints imposed upon it depending on the use case.
User-imposed constraints include the choice of \gls{FPGA} and platform, desired classification throughput in \gls{FPS} and clock frequency.
Simultaneously, the \gls{BNN} topology constrains how the compute resources must be allocated to obtain an efficient heterogeneous streaming architecture.
\OurScheme{} offers parameterizable building blocks and a way of controlling the classification throughput, as described in Sections \ref{sec:BuildingBlocks} and \ref{sec:Folding}.
To achieve portability, we chose a commercial high level synthesis tool, Vivado \gls{HLS}, for the implementation. The tool enables faster development cycles via high-level abstractions, and provides automated pipelining to meet the clock frequency target.

\subsection{\gls{BNN}-specific Operator Optimizations}
\label{sec:Optimizations}
\glspl{BNN} have several properties that enable a more efficient mapping to \glspl{FPGA} without affecting the network accuracy, which we describe in the following subsections.
We assume that the methodology described in \cite{binarynet} is used for training all \glspl{BNN} in this paper, where all \gls{BNN} layers have the following properties (unless otherwise stated):
\begin{itemize}
    \item Using 1-bit values for all input activations, weights and output activations (full binarization), where an unset bit represents -1 and a set bit represents +1.
    \item Batch normalization prior to the activation function.
    \item Using the following activation function: \\
    $\mathrm{Sign}(x) = 
\{+1 \text{~if~} x \geq 0, -1 \text{~if~} x < 0\}$
\end{itemize}

\subsubsection{Popcount for Accumulation}
\label{sec:Popcount}
The regular and value-constrained nature of \gls{BNN} computations enable computing binary dot products with fewer hardware resources.
Let $Y$ be the number of input synapses (or \emph{fan-in}) for a given neuron, with the number of +1-valued synapse inputs denoted as $Y_1$ and -1-valued synapses as $Y_0$.
As there are only two possible values (-1 and +1) for any synapse input, $Y = Y_0 + Y_1$.
Therefore, by counting the number of synapses for only one value, it is possible to infer the summed response for the entire neuron.

The practical consequence for hardware is that the summation of a binary dot product can be implemented by a \emph{popcount} operation that counts the number of set bits instead of accumulation with signed arithmetic.
Our experiments with Vivado HLS indicate that popcount-accumulate requires approximately half the number of LUT and FF resources to implement compared to signed-accumulate.
For instance, with a target $F_{\mathrm{clk}}=200$~MHz, a 128-bit popcount-accumulate requires 376 LUTs and 29 FFs, while a 128-bit bipolar-accumulate requires 759 LUTs and 84 FFs.



\subsubsection{Batchnorm-activation as Threshold}
\label{sec:BatchnormToThres}

All \gls{BNN} layers use batch normalization \cite{batchnorm} on convolutional or fully connected layer outputs, then apply the sign function to determine the output activation. We show how the same output can be computed via thresholding.

Let $a_k$ be the dot product (pre-activation) output of neuron $k$, and $\Theta_k = (\gamma_k, \mu_k, i_k, B_k)$ be the batch normalization parameters learned during training for this neuron.
The output $a^b_k$ is computed as $a^b_k = \mathrm{Sign}(\mathrm{BatchNorm}(a_k, \Theta_k))$, with 
$\mathrm{BatchNorm}(a_k, \Theta_k) = \gamma_k \cdot (a_k - \mu_k) \cdot i_k + B_k$.
Figure \ref{fig:BatchnormAsThres} shows the dot product input vs output activation for three example neurons.
Depending on parameter values, the plot may be shifted towards the left or right, or be flipped horizontally, but a threshold $\tau_k$ for a change in the output activation is always present.
Solving $\mathrm{BatchNorm}(\tau_k, \Theta_k) = 0$ we can deduce that $\tau_k = \mu_k - (B_k/(\gamma_k \cdot i_k))$. 

To make the thresholds compatible with the positive-only operations in Section \ref{sec:Popcount}), the computed threshold is averaged with the neuron fan-in $S$ to obtain $\tau_k^+ = (\tau_k + S)/2$. Observing how neuron C activates with an opposite sign threshold to neurons A and B in Figure \ref{fig:BatchnormAsThres},
all neurons can be made to activate using a greater-than threshold by flipping the signs of a neuron's weights if $\gamma_k \cdot i_k < 0$.

Using these techniques, we can compute the output activation using an unsigned comparison and avoid computing the batch normalized value altogether during inference.
$\tau_k^+$ itself is fixed for a trained network and can be computed from the batchnorm parameters at compile time.
Synthesis reports from Vivado HLS for 16-bit dot product output values indicate that regular batchnorm-and-sign activation requires 2 DSPs, 55 FFs and 40 LUTs, whereas the threshold activation we describe here only requires 6 LUTs.


\begin{figure}
	\centering
	\includegraphics{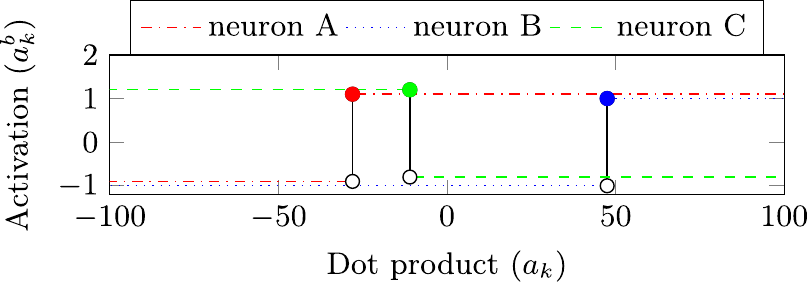}
	\caption{Three examples of binary neuron activations with batch normalization. A slight vertical offset is added for clarity.}
	\label{fig:BatchnormAsThres}
\end{figure}

\subsubsection{Boolean OR for Max-pooling}
\label{sec:BinPooling}
The networks described in \cite{binarynet} perform pooling prior to activations, i.e. pooling is performed on non-binarized numbers, which are then batch normalized and fed into the activation function.
We show that the same layer outputs can be derived by max pooling \emph{after} the activations without having to re-train the network.
Let $a_1, a_2, \ldots a_Y$ be the positive dot product outputs that will be processed by max-pooling.
In accordance with Section \ref{sec:BatchnormToThres}, the output would be computed as $a^b = (\mathrm{Max}(a_1, a_2, \ldots a_Y) > \tau^+)$.
Due to the distributivity of $\mathrm{Max}$, the output will be $\mathnormal{true}$ if \emph{any} of $a_1, a_2, \ldots a_S$ are greater than $\tau^+$.
Therefore, the same result can be computed as $a^b = (a_1 > \tau^+) \vee (a_2 > \tau^+) \ldots \vee (a_Y > \tau^+)$.
As the threshold comparisons are already computed for the activations, max-pooling can be effectively implemented with the Boolean OR-operator.
We note that similar principles apply for min-pooling (as Boolean AND) and average-pooling (as Boolean majority function) as well.


\subsection{FINN Design Flow and Hardware Library}
\label{sec:BuildingBlocks}

\begin{figure}
    \centering
    \includegraphics[height=3.5cm]{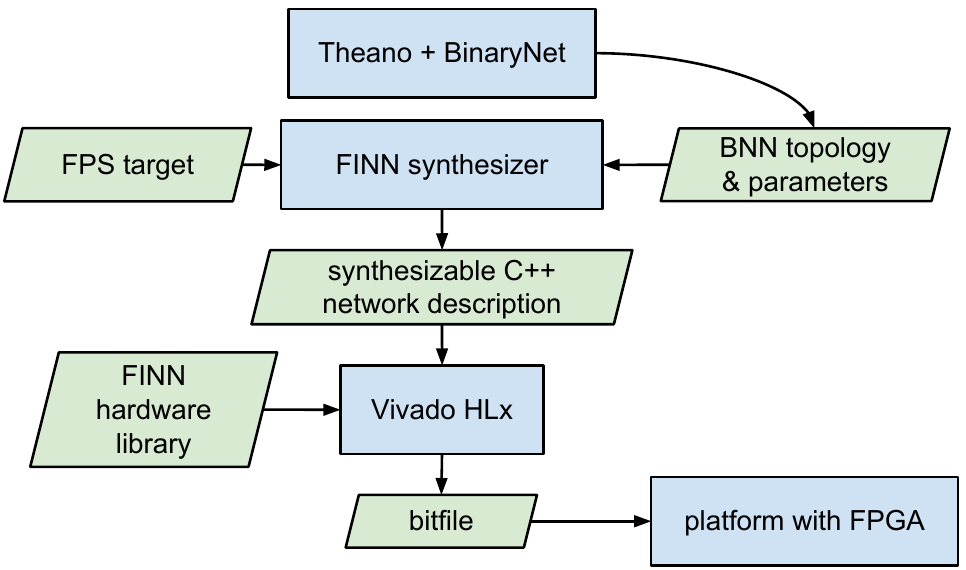}
    \caption{Generating an FPGA accelerator from a trained \gls{BNN}.}
    \label{fig:workflow}
\end{figure}

Figure \ref{fig:workflow} illustrates the design flow for converting a trained BNN into an FPGA accelerator.
The user supplies a \gls{FPS} target alongside a Theano-trained \gls{BNN} to the \OurScheme{} synthesizer.
The synthesizer first determines the folding parameters (Section \ref{sec:Folding}) to meet the \gls{FPS} target and applies the optimizations from Section \ref{sec:Optimizations}, then produces a synthesizable C++ description of a heterogeneous streaming architecture.
The architecture is composed of building blocks from the \OurScheme{} hardware library described in the following subsections.

\subsubsection{The Matrix--Vector--Threshold Unit}
\label{sec:MVTU}
\begin{figure}
    \centering
    \includegraphics[height=3cm]{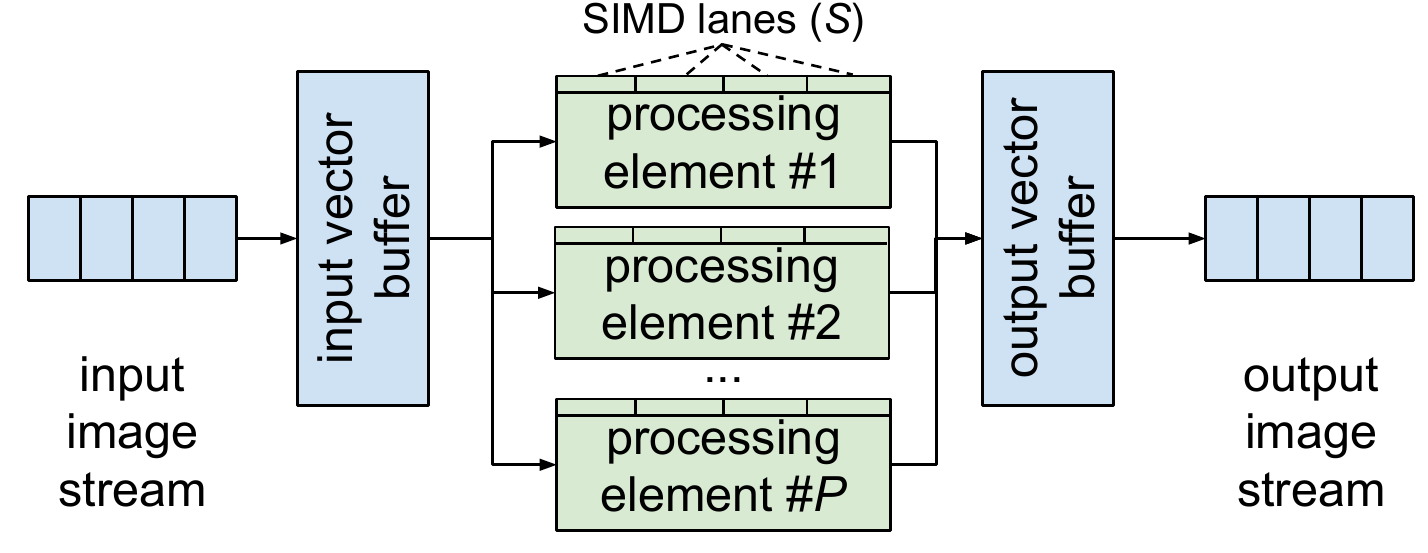}
    \caption{Overview of the MVTU.}
    \label{fig:MVUOverview}
\end{figure}

\begin{figure}
    \centering
    \includegraphics[height=3.2cm]{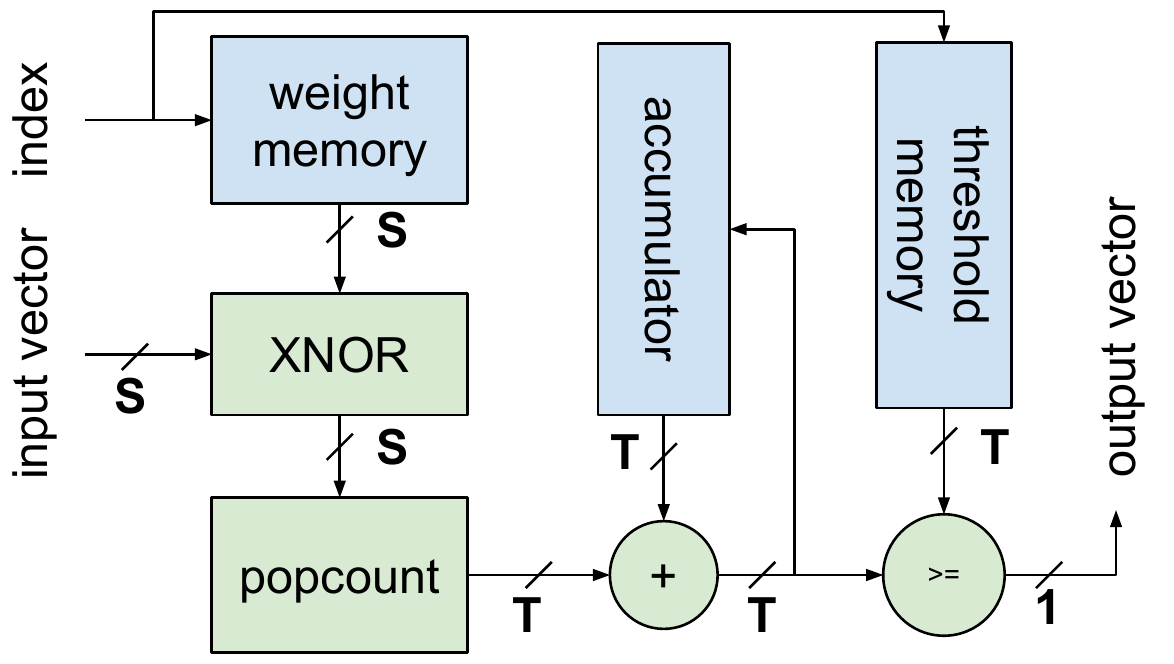}
    \caption{MVTU PE datapath. \textbf{Bold} indicates bitwidth.}
    \label{fig:PEDatapath}
\end{figure}

The \gls{MVU} forms the computational core for our accelerator designs.
The vast majority of compute operations in a \gls{BNN} can be expressed as matrix--vector operations followed by thresholding.
For instance, the pre-activation output $\mathbf{a_N}$ of the fully connected neural network layer at index $N$ is given by matrix-vector product $\mathbf{a_N} = \mathbf{A} \cdot \mathbf{a^b_{N-1}}$ where $\mathbf{A}$ is the synaptic weight matrix and $\mathbf{a^b_{N-1}}$ are the activations from the previous layer.
The post-activation output can then be computed by $\mathbf{a^b_N} = \mathbf{a^N} > \mathbf{\tau^+_N}$, where the thresholds $\mathbf{\tau^+_N}$ are determined as described in Section \ref{sec:BatchnormToThres}.
Convolutions can also be implemented as matrix--vector products, as will be described in Section \ref{sec:ConvImpl}.
As such, the \gls{MVU} implements fully-connected layers as a standalone component, and is also used as part of the convolutional layers.

The overall organization of the \gls{MVU} is shown in Figure \ref{fig:MVUOverview}.
Internally, the \gls{MVU} consists of an input and output buffer, and an array of \glspl{PE} each with a number of SIMD lanes.
The number of \glspl{PE} ($P$) and SIMD lanes ($S$) are configurable to control the throughput as discussed in Section \ref{sec:FoldingMatrixVector}.
The synapse weight matrix to be used is kept in \gls{OCM} distributed between \glspl{PE}, and the input images stream through the \gls{MVU} as each one is multiplied with the matrix.
Each \gls{PE} receives exactly the same control signals and input vector data, but multiply-accumulates the input with a different part of the matrix.
In terms of the taxonomy described in \cite{chen2016eyeriss}, this architecture is both \emph{weight stationary} (since each weight remains local to the \gls{PE}) and \emph{output stationary} (since each popcount computation remains local to the \gls{PE}).

Figure \ref{fig:PEDatapath} shows the datapath of an \gls{MVU} \gls{PE}.
It computes the dot product between the input vector and a row of the synaptic weight matrix and compares the result to a threshold, producing a single-bit output.
The dot product computation itself is a multiply-accumulate operation between the two bit vectors and implemented with an XNOR gate. Following this, the number of set bits in the result is counted (see Section \ref{sec:Popcount}) and added to the accumulator register.
Once the entire dot product is accumulated, it is thresholded.
The accumulator, adder and threshold memory are $T$-bits wide, which can be scaled down to $T=1+\mathrm{log}_{2}(Y)$ for additional resource savings.

Finally, it is worth pointing out that the \gls{MVU} architectural template can also support partial binarization for non-binarized outputs and inputs.
Removing the thresolding stage provides non-binarized outputs, while using regular multiply-add instead of XNOR-popcount can handle non-binarized inputs
These features are used in the first and last layers of networks that process non-binary input images or do not output a one-hot classification vector.

\subsubsection{Convolution: The Sliding Window Unit}
\label{sec:ConvImpl}


Convolutions can be \emph{lowered} to matrix-matrix multiplications \cite{chellapilla2006high}, which is the approach followed in this work.
The weights from the convolution filters are packed into a \emph{filter matrix}, while a sliding window is moved across input images to form an \emph{image matrix}.
These matrices are then multiplied to generate the output images.

\begin{figure}
	\centering
	\begin{subfigure}[t]{0.8\linewidth}
	  \centering
	  \includegraphics[width=\linewidth]{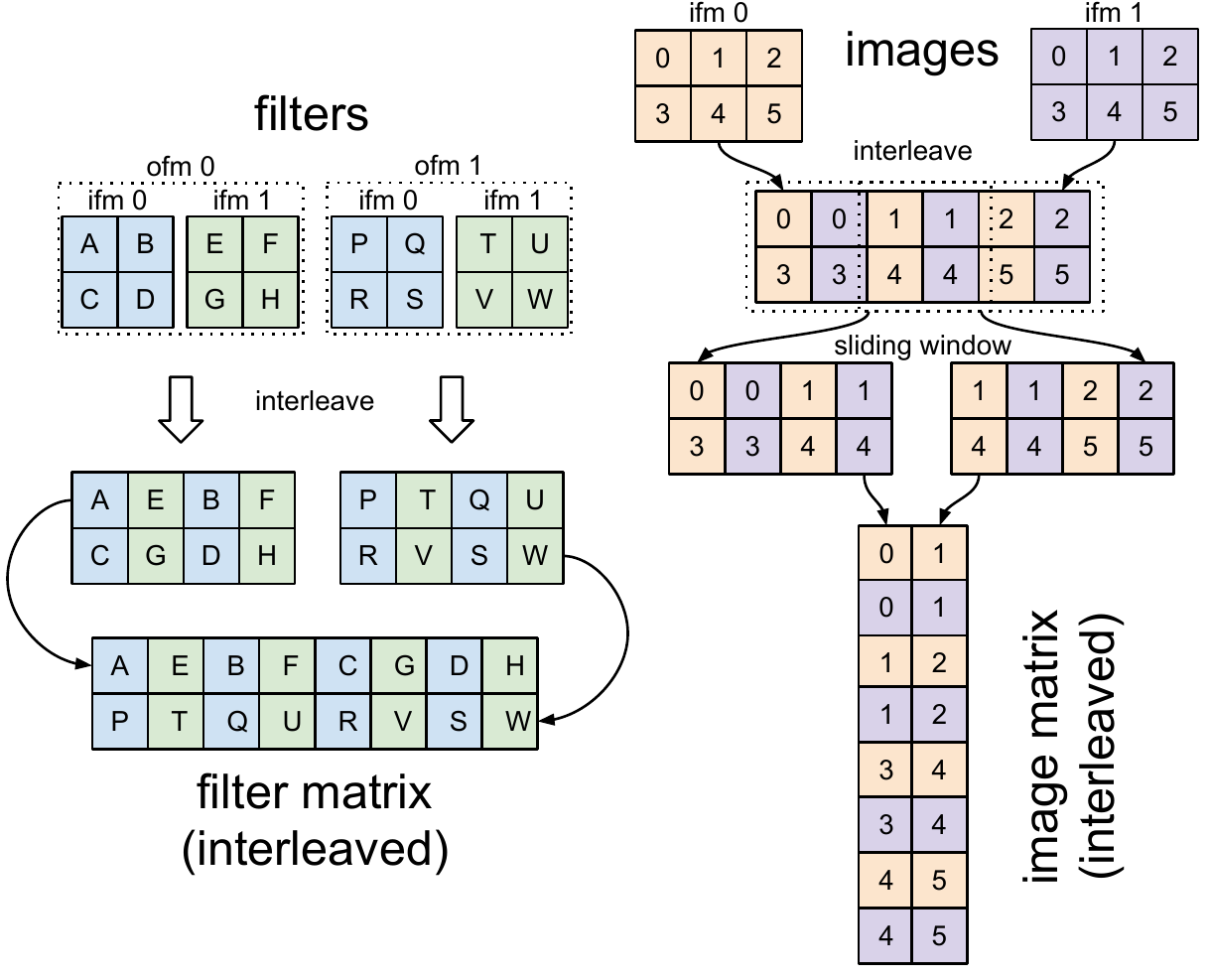}
	  \caption{\small Lowering with interleaved channels.}
	  \label{fig:ConvInterleaving}
	\end{subfigure}

	\begin{subfigure}[t]{0.8\linewidth}
		\centering
		\includegraphics[width=\linewidth]{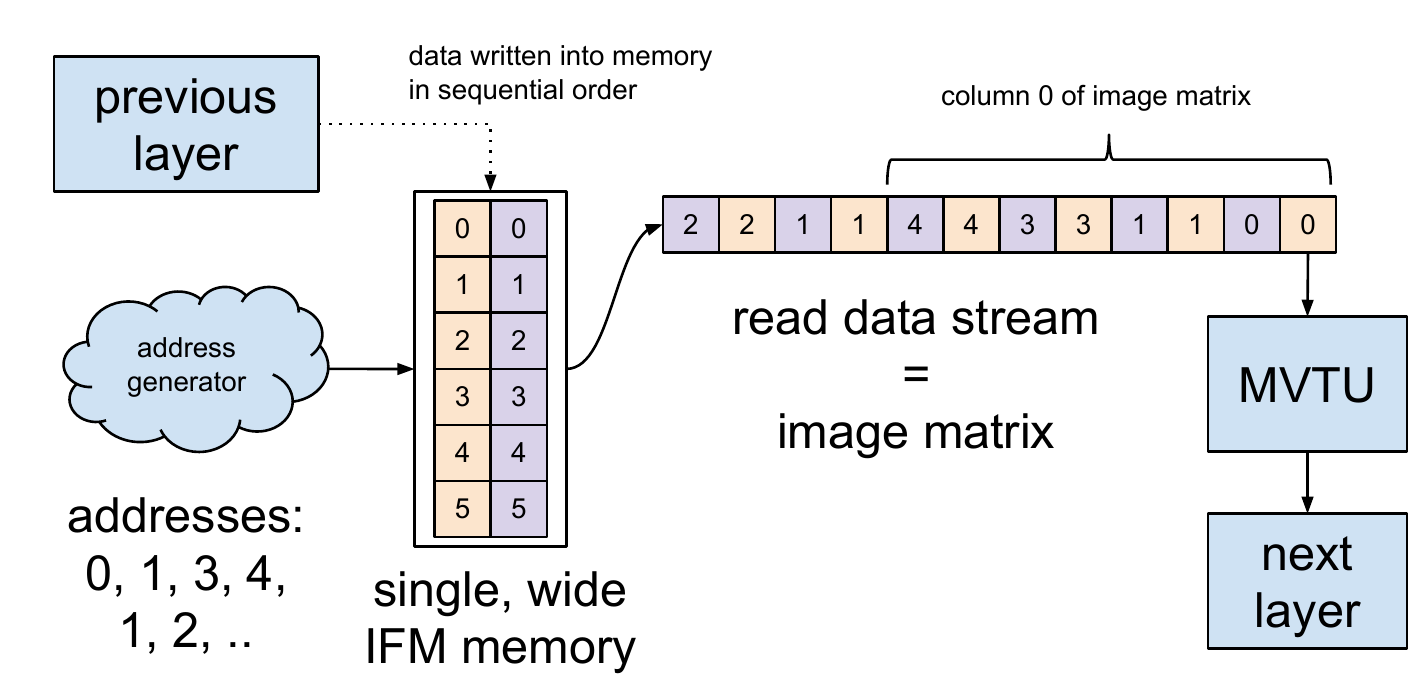}
		\caption{\small SWU operation.}
		\label{fig:SWUImpl}
	\end{subfigure}

	\caption{Convolution using interleaved channels.}
\end{figure}

The convolutional layer consists of a \gls{SWU}, which generates the image matrix from incoming feature maps, and a \gls{MVU} that actually computes the matrix--matrix product using a different column vector from the image matrix each time.
In order to better cater for the SIMD parallelism of the \gls{MVU} and minimize buffering requirements, we \emph{interleave} the feature maps such that each pixel contains all the \gls{IFM} channel data for that position, as illustrated in Figure \ref{fig:ConvInterleaving}.
Since the dot product to compute a \gls{OFM} pixel includes all \glspl{IFM} pixels at a certain sliding window location, those \gls{IFM} pixels can be processed in any order owing to the commutative property of addition.
Note that interleaving the filter matrix has no additional cost since it is done offline, and interleaving the input image can be done on-the-fly in the FPGA.
Storing the pixels in this fashion allows us to implement the \gls{SWU} with a single wide \gls{OCM} instead of multiple narrow \glspl{OCM}, and also enables the output of the \gls{MVU} to be directly fed to the next layer without any transposition.
As illustrated in Figure \ref{fig:SWUImpl}, the incoming \gls{IFM} data is simply stored at sequential addresses in a buffer, then the memory locations corresponding to each sliding window are read out to produce the image matrix.


Although not required by any of the networks described in this work, the \gls{SWU} also pads the images if necessary.
One interesting observation is that with the bipolar number representation used in this work, there is no number corresponding to zero.
Therefore, in order to maintain a true binary datapath for activations, images must be padded with our representation or either a 1 or a -1.
Future work will look into what impact this has on the accuracy of trained networks, but early experiments suggest that there is very little difference in accuracy, with respect to \cite{binarynet}.

\subsubsection{The Pooling Unit}
The \gls{PU} implements max-pooling as described in Section \ref{sec:BinPooling}.
To implement $k \times k$ max-pooling on a $D_H \times D_W$ binary image of $C$ channels, the \gls{PU} contains $C \cdot k$ line buffers of $D_W$ bits each.
As with the rest of our component library, the \gls{PU} operates in a streaming fashion.
The input image is gradually streamed into the line buffers.
When at least $k$ rows of the image have arrived, each $k$ consecutive bits of the line buffer are OR'ed together to produce horizontal subsampling for each channel.
These are then OR'ed together with the other line buffers to produce vertical subsampling, the results are streamed out, and the oldest line buffers are refilled with the next row of pixels.

\subsection{Folding}
\label{sec:Folding}
In terms of the \gls{MVU} description given in Section \ref{sec:MVTU}, each \gls{PE} corresponds to a \emph{hardware neuron}, while each SIMD lane acts as a \emph{hardware synapse}.
If we were to dimension each \gls{MVU} in a network with a number of hardware neurons and synapses equal to the number of neurons and synapses in a \gls{BNN} layer, this would result in a fully parallel neural network that could classify images at the clock rate.
However, the amount of hardware resources on an \gls{FPGA} is limited, and it is necessary to time-multiplex (or \emph{fold}) the \gls{BNN} onto fewer hardware synapses and neurons.
We now describe how the folding is performed subject to user constraints.

The work by Venieris et al. \cite{venieris2016fpgaconvnet} describes a method for folding neural networks expressed as streaming dataflow graphs, with focus on formalizing the folding and design space exploration.
In this work, we consider a simpler variant that only controls the folding of matrix--vector products to achieve a given \gls{FPS} requirement set by the user, and focus on \emph{how} the folding is implemented in terms of the workload mapping.
As almost all computations in \glspl{BNN} are expressed as matrix--vector multiplications, implementing folding for matrix--vector multiplication already enables a great degree of control over the system throughput.
Folding directly affects the resource and power consumption of the final system as well, which we explore in Section \ref{sec:results}.

\subsubsection{Folding Matrix--Vector Products}
\label{sec:FoldingMatrixVector}
\begin{figure}
    \centering
    \includegraphics[width=0.8\linewidth]{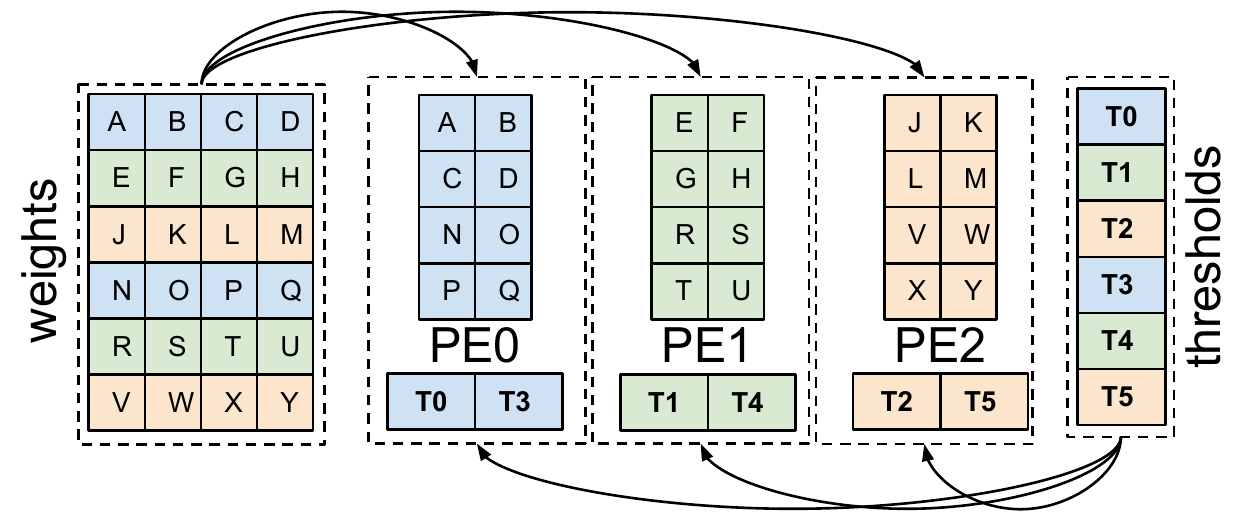}
    \caption{Neuron and synapse folding for \gls{MVU}.}
    \label{fig:MVUMapping}
\end{figure}

Folding matrix--vector products is achieved by controlling two parameters of the \gls{MVU}: $P$ the number of \glspl{PE}, and $S$ the number of SIMD lanes per \gls{PE}.
These determine how the matrix is partitioned between the \glspl{PE}.
A $P$-high, $S$-wide tile of the matrix is processed at a time, with each row in the tile mapped to a different PE, and each column to a different SIMD lane.
For a $X \times Y$ matrix, we refer to $F^n = X / P$ as the \emph{neuron fold} and $F^s = Y / S$ as the \emph{synapse fold}.
The \emph{total fold} $F$ is then obtained as $F = F^n \cdot F^s$, which is also the number of cycles required to complete one matrix--vector multiply.
Note that $F^n$ and $F^s$ should be integers to avoid padding the weight matrix.
As an example, Figure \ref{fig:MVUMapping} shows how a $6\times4$  weight matrix is partitioned between three \glspl{PE} with two SIMD lanes each.
Here, each matrix-vector multiply will take $F^n \cdot F^s = (6 / 3) \cdot (4 / 2) = 4$ cycles.

The same principle applies for convolutional layers, but these always have an inherent amount of folding due to our current matrix--matrix product as multiple matrix--vector products implementation.
For convolutional layers, the total fold is $F = F^m \cdot F^n \cdot F^s$, where 
$F^m$ is a network-dependent constant due to multiple matrix-vector products, and is equal to the number of output pixels from the convolution.

\subsubsection{Determining $F^n$ and $F^s$}
Avoiding the \dblquotes{one-size-fits-all} inefficiencies requires tailoring each \gls{MVU}'s compute resources to layer requirements.
The guiding principle here is \emph{rate-balancing} the heterogeneous streaming architecture: the slowest layer (with $II_{\mathrm{max}}$) will determine the overall throughput, so each layer should use a roughly equal number of cycles to process one image.
As this is a streaming system, the classification throughput $\mathrm{FPS}$ will be approximately $\frac{F_{\mathrm{clk}}}{II_{\mathrm{max}}}$, where $F_{\mathrm{clk}}$ is the clock frequency.
For a fully-connected layer, the total fold $F$ is equal to the \gls{II}.
Therefore, balancing a fully-connected \gls{BNN} can be achieved by using $F^n$ and $F^s$ such that $F^n \cdot F^s = \frac{F_{\mathrm{clk}}}{\mathrm{FPS}}$ for each layer.
Depending on the \gls{BNN} and the \gls{FPS} requirements, the number of memory channels or sliding window generation may constitute bottlenecks.
For such cases, we match the throughput of all other layers to the bottleneck in order not to waste resources.

\section{Evaluation}
\label{sec:results}

\subsection{Experimental Setup}

To evaluate \OurScheme{}, we created a number of prototypes that accelerate \glspl{BNN} inference on the MNIST \cite{mnist} ($28 \times 28$ handwritten digits), CIFAR-10 \cite{cifar10} ($32 \times 32$ color images in 10 categories) and cropped SVHN \cite{svhn} ($32 \times 32$ images of Street View House Numbers) datasets.
Each prototype combines a \gls{BNN} topology with a different use case scenario.
We consider three different \gls{BNN} topologies for classifying the datasets as follows:

\begin{itemize}
	\item \textbf{SFC} and \textbf{LFC} are three-layer fully connected network topologies for classifying the MNIST dataset, with different numbers of neurons to demonstrate accuracy-computation tradeoffs (Section \ref{subsec:accuracy}). SFC contains 256 neurons per layer and achieves 95.83\% accuracy, while LFC has 1024 neurons per layer and achieves 98.4\% accuracy. These networks accept 28x28 binary images and output a 10-bit one-hot vector indicating the digit.

    \item \textbf{CNV} is a convolutional network topology inspired by BinaryNet \cite{binarynet} and VGG-16 \cite{simonyan2014very}. It contains a succession of (3x3 convolution, 3x3 convolution, 2x2 maxpool) layers repeated three times with 64-128-256 channels, followed by two fully connected layers of 512 neurons each. We use this topology for classifying both the CIFAR-10 (with 80.1\% accuracy) and SVHN (with 94.9\% accuracy) datasets, with different weights and thresholds. Note that the inputs to the first layer and the outputs from the last layer are not binarized; CNV accepts 32x32 images with 24 bits/pixel, and returns a 10-element vector of 16-bit values as the result.
\end{itemize}

To further demonstrate the flexibility of the framework, we consider two usage scenarios for each \gls{BNN} topology to guide the choice of parametrization:

\begin{itemize}
    \item \textbf{max} is the maximum performance scenario where it is desirable to reach the peak \gls{FPS} permitted by the platform, topology and \OurScheme{}'s architecture.
    \item \textbf{fix} represents a scenario with a fixed \gls{FPS} requirement, which is often determined by an I/O device for real life applications. For instance, consider a $640 \times 480$ video stream at 30~\gls{FPS}, which is to be chopped up into $32 \times 32$ tiles for neural network inference. Handling this task with real-time performance would require a \gls{BNN} inference rate of 9000~\gls{FPS}, which we set as the requirement for this usage scenario.
\end{itemize}

We use shortened names to refer to the prototypes, e.g. CNV-fix refers to the prototype that implements the \textbf{CNV} topology for the \textbf{fix} usage scenario.
For each prototype, the folding factors (Section \ref{sec:Folding}) were determined to meet the requirements of its usage scenario, and the \OurScheme{} design flow (Section \ref{sec:BuildingBlocks}) was followed to generate the hardware accelerator.
Vivado HLS and Vivado version 2016.3 were used for the bitfile synthesis.
A target clock frequency of 200~MHz was used for both Vivado HLS and Vivado, and to run the resulting accelerator unless otherwise stated.
The salient properties of the topologies and folding factors for the prototypes are summarized in Table \ref{tab:SummaryWorkloads}.

All prototypes have been implemented on the Xilinx Zynq-7000 All Programmable SoC ZC706 Evaluation Kit running Ubuntu 15.04.
The board contains a Zynq Z7045 SoC with dual ARM Cortex-A9 cores and FPGA fabric with 218600 LUTs and 545 BRAMs.
The host code runs on the Cortex-A9 cores of the Zynq.
It initializes 10000 images with test data in the Zynq's shared DRAM, launches and times the accelerator execution to measure classification throughput, then measures accuracy by comparing against the correct classifications.
Two power measurements \PowerFPGA{} and \PowerWall{} are provided for each experiment; \PowerFPGA{} using the PMBus interface to monitor the \gls{FPGA} power supply rails, and \PowerWall{} using a wall power meter for the total board power consumption.
The measurements are averaged over a period of 10 seconds while the accelerator is running.

\begin{table}
	\centering
	\caption{Summary of workloads.}
	\scriptsize
	\begin{tabularx}{\linewidth}{l|cccc}
\toprule
Topology & Params & Ops & Off-chip I/O & Op.Int.\\
 & (Mbits) & (M) & (B) &  (Ops/B) \\
\midrule
SFC & 0.3 & 0.6 & 112 & 5970 \\
LFC & 2.9 & 5.8 & 112 & 51968 \\
CNV & 1.5 & 112.5 & 3092 & 36400 \\
\midrule
Prototype & \multicolumn{4}{c}{Per-Layer Total Fold $(F)$} \\
\midrule
SFC-max & \multicolumn{4}{c}{13, 16, 16, 16} \\
SFC-fix & \multicolumn{4}{c}{12544, 16384, 16384, 2560} \\
LFC-max & \multicolumn{4}{c}{104, 128, 128, 128} \\
LFC-fix & \multicolumn{4}{c}{13312, 16384, 16384, 10240} \\
CNV-max & \multicolumn{4}{c}{8100, 7056, 5184, 7200, 5184, 4608, 8192, 8192, 1280} \\
CNV-fix & \multicolumn{4}{c}{\tiny 16200, 14112, 10368, 14400, 10368, 9216, 16384, 16384, 1280} \\
\bottomrule
\end{tabularx} 
	\label{tab:SummaryWorkloads}
\end{table}
\subsection{Results}

\begin{table}
	\centering
	\caption{Summary of results from \OurScheme{} 200~MHz prototypes.}
	\scriptsize
	\begin{tabularx}{\linewidth}{l|XXXXXX}
\toprule
Name & Thr.put & Latency & LUT & BRAM & \PowerFPGA{} & \PowerWall{} \\
 & (FPS) & (\textmu s) &  &  & (W) & (W) \\
\midrule
SFC-max & 12361~k & 0.31 & 91131 & 4.5 & 7.3 & 21.2 \\ 
LFC-max & 1561~k & 2.44 & 82988 & 396 & 8.8 & 22.6 \\ 
CNV-max & 21.9~k & 283 & 46253 & 186 & 3.6 & 11.7 \\ 
SFC-fix & 12.2~k & 240 & 5155 & 16 & 0.4 & 8.1 \\ 
LFC-fix & 12.2~k & 282 & 5636 & 114.5 & 0.8 & 7.9 \\ 
CNV-fix & 11.6~k & 550 & 29274 & 152.5 & 2.3 & 10 \\ 
\bottomrule
\end{tabularx} 
	\label{tab:SummaryResults}
\end{table}

Table \ref{tab:SummaryResults} provides an overview of the experimental results, in terms of classification throughput, latency to classify one image, \gls{FPGA} resource usage and power.
The \textbf{max} scenario results are perhaps the best summary of the potential of \glspl{BNN} on \glspl{FPGA}, with SFC-max achieving 12.3 million classifications per second at 0.31 \textmu s latency while drawing less than 22~W total power.
All \textbf{fix} results meet and exceed the 9000~FPS requirement by 30\% due to folding factors being integers, though lower throughput and power could have been achieved by using a slower clock.
We focus on particular aspects of the results in the following subsections.

\subsubsection{Maximum Throughput and Bottlenecks}
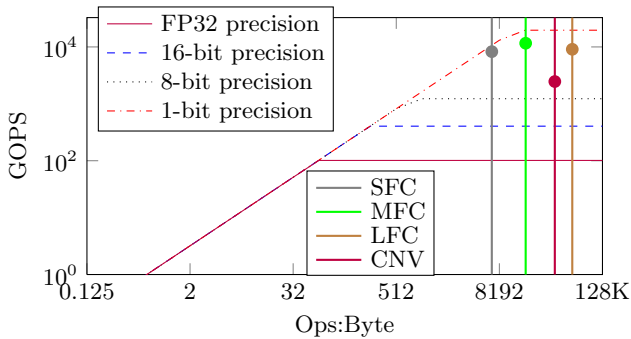
\begin{figure}
\begin{tikzpicture}
\begin{loglogaxis}[
	width=\linewidth,height=5cm,
    xlabel={Ops:Byte},
    ylabel={GOPS},
    log basis x=2,
    xtick={0.125, 2, 32, 512, 8192, 131072},
    xticklabels={0.125, 2, 32, 512, 8192, 128K},
    xmin=0.125, xmax=131072,
    ymin=1, ymax=32768,
    legend pos=north west
]
 
\addplot[
    dashed,
    color=blue,
    mark=none,
    ]
    table [x=OperationalIntensity, y=ZC706-HP, col sep=comma] {csv-graph/Roofline-updated.csv};
    \label{16-bit precision};
\addplot[
    dotted, 
    color=black,
    mark=none,
    ]
    table [x=OperationalIntensity, y=ZC706-QP, col sep=comma] {csv-graph/Roofline-updated.csv};
    \label{8-bit precision};
\addplot[
    dashdotted,
    name path global=1bit,
    color=red,
    mark=none,
    ]
    table [x=OperationalIntensity, y=ZC706-1b, col sep=comma] {csv-graph/Roofline-updated.csv};
    \label{1-bit precision}; 
\addplot[
    name path global=1bit,
    color=purple,
    mark=none,
    ]
    table [x=OperationalIntensity, y=ZC706-FP, col sep=comma] {csv-graph/Roofline-updated.csv};
    \label{FP precision}; 
\addplot [color=brown, thick, name path global=LFC] coordinates {
		(58204,1)
		(58204,32768)};
    \label{LFC};
\addplot [color=brown, thick, mark=*] coordinates {
		(58204,9086)};
\addplot [color=gray, thick, name path global=MFC] coordinates {
		(6686,1)
		(6686,32768)};
    \label{SFC};
\addplot [color=gray, thick, mark=*] coordinates {
		(6686,8265)};
\addplot [color=green, thick, name path global=SFC] coordinates {
		(16621,1)
		(16621,32768)};
    \label{MFC};
\addplot [color=green, thick, mark=*] coordinates {
		(16621,11612)};
\addplot [color=purple, thick, name path global=CNV] coordinates {
		(36400,1)
		(36400,32768)};
    \label{CNV};
\addplot [color=purple, thick, mark=*] coordinates {
		(36400,2465)};

\end{loglogaxis}

\node [draw,fill=white] at (rel axis cs: 0.4,0.8) {\shortstack[l]{
\ref{FP precision} FP32 precision \\
\ref{16-bit precision} 16-bit precision \\
\ref{8-bit precision} 8-bit precision \\
\ref{1-bit precision} 1-bit precision}};

\node [draw,fill=white] at (rel axis cs: 0.7,0.2) {\shortstack[l]{
\ref{SFC} SFC \\
\ref{MFC} MFC \\
\ref{LFC} LFC \\
\ref{CNV} CNV 
}};
\end{tikzpicture}
\caption{ZC706 roofline with topologies and \textbf{max}-datapoints.}
\label{fig:roofline_zc706}
\end{figure}
To assess the quality of results for the \textbf{max} scenarios, we compare the achieved performance (XNOR--popcount operations per second) with the peak throughput in \gls{TOPS} indicated by the roofline model.
Figure \ref{fig:roofline_zc706} presents a roofline model (Section \ref{subsec:back-rooflines}) for the ZC706, assuming 90\% LUT utilization, 200~MHz clock frequency and 1.6 GB/s of DRAM bandwidth.
The vertical lines show the arithmetic intensities for the topologies, and the actual operations per second values from corresponding prototypes with \textbf{max} usage scenarios are indicated as points on those lines.
All \textbf{max} prototypes achieve performance in the \gls{TOPS} range, but are bottlenecked due to different factors.
CNV-max achieves 2.5 \gls{TOPS} and is \emph{architecture-bound}. The current \gls{SWU} design does not scale as well as the \gls{MVU} and constitutes a bottleneck, which will be addressed in future work.
Despite its higher complexity, observe that CNV-max actually requires $\approxtilde{2}\times$ fewer LUTs than SFC-max since the folding parameters for CNV-max are chosen in accordance with the maximum performance dictated by the bottleneck.
SFC-max achieves 8.2 \gls{TOPS} and is \emph{memory-bound}. Observe that the SFC arithmetic intensity line intersects the memory-bound (sloped) part of the roofline, thus the performance cannot be scaled up without adding more DRAM memory bandwidth.
LFC-max achieves 9.1 \gls{TOPS}, which is 46\% of the roofline, and is \emph{resource-bound}. 
As folding factors are integers, the smallest increment is $2\times$ which roughly doubles the resource cost. 
The FPGA has enough LUTs but not enough BRAMs to accommodate doubled resource cost, thus leaving \approxtilde{30\%} of BRAMs unused.
A 3x512-neuron fully connected topology, labeled MFC in Figure \ref{fig:roofline_zc706}, was able to achieve 11.6 \gls{TOPS} and 6238 kFPS with 95\% of the device BRAMs.

\subsubsection{Energy Efficiency}
\begin{figure}
\begin{tikzpicture}
\begin{semilogyaxis}[
			width=\linewidth, height=3.2cm,
			xtick={1,2,3,4,5,6,7,8,9,10,11},
			legend style={font=\scriptsize},
			legend pos={north east},
			legend columns=2,
			ylabel={FPS/W},
			xticklabel style={rotate=15,anchor=east, font=\scriptsize},
			ybar, bar width=4pt,
			symbolic x coords={SFC-max, LFC-max, CNV-max, SFC-fix, SFC-smax, LFC-fix, CNV-fix},
			xtick={SFC-max, LFC-max, CNV-max, SFC-fix, SFC-smax, LFC-fix, CNV-fix},
			]
			
\addplot+[ybar][domain=0:1000] coordinates {					
(SFC-max, 583066.0377)
(LFC-max, 69070.79646)
(CNV-max, 1872.307692)
(SFC-fix, 1506.740741)
(LFC-fix, 1544.886076)
(CNV-fix, 1160.99)
(SFC-smax, 2247.285714)

};					
\addlegendentry{wall}					
					
\addplot+[ybar] coordinates {					
(SFC-max, 1693287.671)
(LFC-max, 177386.3636)
(CNV-max, 6085)
(SFC-fix, 28382.7907)
(LFC-fix, 15255.75)
(CNV-fix, 5047.782609)
(SFC-smax, 78655)
};					
\addlegendentry{chip}										

\end{semilogyaxis}

\end{tikzpicture}
\caption{Prototype energy efficiency.}
\label{fig:EnergyEff}
\end{figure}
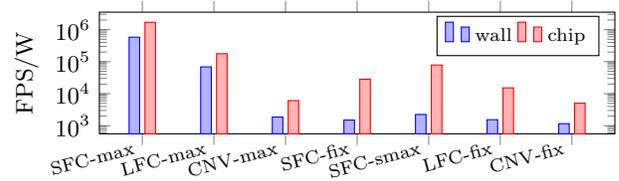
It is desirable to minimize the energy spent per image classification, which corresponds to maximizing \gls{FPS} per Watt when many images are to be classified.
To help evaluate the energy efficiency, Figure \ref{fig:EnergyEff} plots the achieved FPS per Watt for the prototypes for both the wall power and FPGA power readings.
In general, we see that the higher FPS prototypes have better energy efficiency, with SFC-max offering 583066 FPS per W of total power and outperforming all other prototypes by at least an order of magnitude.
It is also worth noting that the board's idle power consumption is about 7~W, which forms a lower bound on all wall power measurements, and could be improved by e.g. using LPDDR memory.

To maximize energy efficiency with a fixed target FPS, is it better to use a highly parallel design at low clock frequency, or a less parallel design at high clock frequency?
We ran an additional experiment to investigate this question by slowing down the SFC-max prototype to meet the \textbf{fix} FPS requirement of 9000 FPS.
By clocking it at 250~kHz, we obtained a classification throughput of 15731~FPS with 0.2~W of FPGA power.
The result is labeled SFC-smax in Figure \ref{fig:EnergyEff}, and is over $2\times$ more energy efficient than SFC-fix.
This suggests that a high degree of parallelism benefits energy efficiency as long as the \gls{FPGA} resources are available.

\subsubsection{Resource Efficiency}
We consider two aspects of resource efficiency for \OurScheme{}: how efficiently the compute units are used during runtime (\emph{runtime efficiency}), and how efficiently FPGA resources are turned into compute units (\emph{mapping efficiency}).

To assess runtime efficiency, we divide the FPS-based (actual) operations per cycle ($\frac{\mathrm{FPS} \cdot \mathrm{Ops}}{F_{\mathrm{clk}}}$) by the (peak) number of synaptic operations per cycle from the design ($\sum{2 \cdot P \cdot S}$).
The prototypes exhibit good runtime efficiency, with \approxtilde{70\%} for \textbf{CNV}, \approxtilde{80\%} for \textbf{SFC} and \approxtilde{90\%} for \textbf{LFC}.
The efficiency can be increased further by fine-tuning the folding factors between different layers.

\begin{figure}
\begin{tikzpicture}

\begin{axis}[
			axis y line*=left,
			width=\linewidth, height=3cm,
			xlabel={PE count},
			ylabel={LUT/Ops/cycle},
			]
			
\addplot+[solid] coordinates {
(256, 2.188293457)
(128, 2.566101074)
(64, 1.658447266)
(32, 1.83203125)
(8, 2.58984375)
(2, 5.015625)
(1, 8.3125)
}; \label{luteff}
\addlegendentry{LUT efficiency}
\end{axis}

\begin{axis}[
			axis y line*=right,
			axis x line = none,
			width=\linewidth, height=3cm,
			ylabel={BRAMs},
			]

\addlegendimage{/pgfplots/refstyle=luteff}\addlegendentry{LUT efficiency}			
\addplot+[solid,mark=o,red] coordinates {
(256, 0)
(128, 0)
(64, 64)
(32, 32)
(8, 8)
(2, 3)
(1, 2.5)
};					
\addlegendentry{BRAM usage}

\end{axis}

\end{tikzpicture}
\caption{Mapping resource efficiency.}
\label{fig:MappingResEff}
\end{figure}
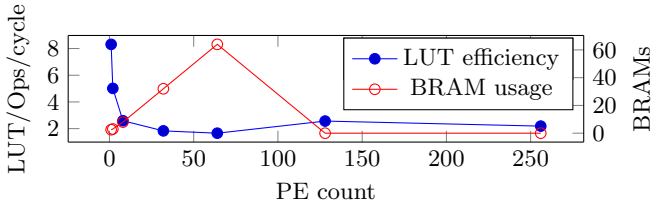

Evaluating the mapping efficiency directly on the prototypes loses some insight, since \textbf{CNV} uses LUTs on \gls{SWU} and \gls{PU}, while fully-connected topologies do not.
Instead, for a single $256 \times 256$ fully-connected layer, we fix $S=64$ and vary $P$, and plot the LUTs per synaptic operation in Figure \ref{fig:MappingResEff}, which should be minimized to maximize efficiency.
The LUTs per operation decreases with higher $P$ since the fixed-size control logic is amortized between more \glspl{PE} and reaches a minimum of 1.83 for $P=64$, but increases again for $P>64$.
To understand why, we also plot the number of BRAMs used in the same figure.
Although all designs have the same number of \gls{BNN} parameters, the number of BRAMs increases with $P$ since each \gls{PE} needs its own weight and threshold memories.
This also means a significant part of the BRAM storage capacity is unused for $1 < P \leq 64$, since the same amount of network parameters is divided between a greater number of memories.
This is also visible for SFC-fix and SFC-max, which use the same network parameters, but have almost $10\times$ difference in the number of BRAMs used (15.5 vs 130.5) since SFC-max has more compute elements working in parallel.
Here, with $P>64$, so little of each BRAM is used that Vivado HLS implements the weight and threshold memories using LUTs, which causes the LUTs per operation to increase.
Thus, the depth and number of BRAMs, and the LUT-to-BRAM ratio of the FPGA plays a key role in determining how well the resources will be utilized by a \gls{BNN}.
For instance, on another FPGA with the same amount of LUTs but twice the number of half-depth BRAMs, LFC-max could achieve $2\times$ throughput.

\subsection{Comparison to prior work}
From an application perspective, we suggest that the current best way to compare different platforms is to simply compare their accuracy, FPS and power consumption when working on the same benchmark datasets (MNIST, CIFAR-10 and SVHN).
This comparison is provided in Table~\ref{tab:comparison}, and is divided into three sections: our results, prior work on low-precision ($<4$ bits) networks, and prior work with higher-precision ($>4$ bits) networks.

When it comes to pure image throughput, our designs outperform all others.
For the MNIST dataset, we achieve an FPS which is over 48/6$\times$ over the nearest highest throughput design~\cite{alemdar2016ternary} for our SFC-max/LFC-max designs respectively.
While our SFC-max design has lower accuracy than the networks implemented by Alemdar~et~al.~\cite{alemdar2016ternary} our LFC-max design outperforms their nearest accuracy design by over 6/1.9$\times$ for throughput and FPS/W respectively. 
For other datasets, our CNV-\textit{max} design outperforms TrueNorth~\cite{esser2016convolutional} for FPS by over 17/8$\times$ for CIFAR-10 / SVHN datasets respectively, while achieving 9.44$\times$ higher throughput than the design by Ovtcharov~et~al.~\cite{ovtcharov2015accelerating}, and $2.2\times$ over the fastest results reported by Hegde~et~al.~\cite{caffepresso}.
Our prototypes have classification accuracy within 3\% of the other low-precision works, and could have been improved by using larger \glspl{BNN}.

\begin{table*}
	\centering
	\caption{Comparison to prior work. Metrics not reported by prior work are indicated by dashes (-), and our estimates by \approxtilde{ }.}
	\tiny
	\resizebox{\linewidth}{!}{
\begin{tabularx}{\linewidth}{l|rrrrrrrrrr}
\toprule
Name & Dataset & Platform & Precision & Err. (\%) & kFPS & \PowerFPGA{} (W) & \PowerWall{} (W) & kFPS/\PowerFPGA{} & kFPS/\PowerWall{} & GOPS \\
\midrule
SFC-max & MNIST & ZC706 & 1 & 4.17 & 12,361 & 7.3 & 21.2 & 1693.29 & 583.07 & 8,265.45 \\
LFC-max & MNIST & ZC706 & 1 & 1.60 & 1,561 & 8.8 & 22.6 & 177.39 & 69.07 & 9,085.67 \\
MFC-max & MNIST & ZC706 & 1 & 2.31 & 6,238 & 11.3 & 28.5 & 552 & 218.8 & 11,612.86 \\
CNV-max & CIFAR-10 & ZC706 & 1 & 19.90 & 21.9 & 3.6 & 11.7 & 6.08 & 1.87 & 2,465.5 \\
CNV-max & SVHN & ZC706 & 1 & 5.10 & 21.9 & 3.6 & 11.7 & 6.08 & 1.87 & 2,465.5 \\
\midrule
Alemdar~et~al.~\cite{alemdar2016ternary} & MNIST & Kintex-7 160T & 2 & 2.24 & 255.10 & 0.32 & - & 806.45 & - & \approxtilde{96.68} \\
Alemdar~et~al.~\cite{alemdar2016ternary} & MNIST & Kintex-7 160T & 2 & 1.71 & 255.10 & 1.84 & - & 138.50 & - & \approxtilde{448.47} \\
Alemdar~et~al.~\cite{alemdar2016ternary} & MNIST & Kintex-7 160T & 2 & 1.67 & 255.10 & 2.76 & - & 92.59 &  - & \approxtilde{864.03} \\
Park and Sung~\cite{park2016fpga} & MNIST & ZC706 & 3 & - & 70 & 4.98 & - & 14.06 & - & \approxtilde{210} \\
TrueNorth~\cite{esser2016convolutional} & CIFAR-10 & TrueNorth & 1 & 16.59 & 1.249 & 0.2044 & - & 6.11 & - & - \\
TrueNorth~\cite{esser2016convolutional} & SVHN & TrueNorth & 1 & 3.34 & 2.526 & 0.2565 & - & 9.85 & - & - \\
\midrule
CaffePresso~\cite{caffepresso} & MNIST & Keystone-II & 16  & - & 5 & - & 14 & - & 0.357 & 44.82 \\
CaffePresso~\cite{caffepresso} & CIFAR-10 & Keystone-II & 16  & - & 10 & - & 14 & - & 0.714 & 146.14 \\
CaffePresso~\cite{caffepresso} & MNIST & Parallella & 32 & - & 0.64 & - & 5 & - & 0.129 & 5.78 \\
CaffePresso~\cite{caffepresso} & CIFAR-10 & Parallella & 32  & - & 0.1 & - & 5 & - & 0.019 & 1.40 \\
Ovtcharov~et~al.~\cite{ovtcharov2015accelerating} & CIFAR-10 & Stratix V D5 & 32 & \approxtilde{11-26} & 2.32 & - & 25 & - & 0.093 & - \\
\bottomrule
\end{tabularx}
}
	\label{tab:comparison}
\end{table*}


\section{Conclusion}
\gls{BNN} for image classification have recently been proposed and this work demonstrates their promise for high performance implementation. 
They are particularly well-suited for FPGA implementations as parameters can be fit entirely in OCM and arithmetic is simplified, enabling high computational performance.
The novel parameterizable dataflow architecture and optimizations presented enable unprecedented classification rates, minimal power consumption and latency, while offering the flexibility and scalability required for accelerating larger and more complex networks. 
We hence believe that this technology is eminently suitable for embedded applications requiring real-time response, including surveillance, robotics and augmented reality. Future work will focus on providing support for non-binary low precision, implementing larger networks like AlexNet, higher performance convolutions, and a more thorough design space exploration. Finally, \OurScheme{} assumes that all \gls{BNN} parameters can fit into the available \gls{OCM} of a single \gls{FPGA}. Supporting external memory, multi-\glspl{FPGA} implementations and reconfiguration~\cite{venieris2016fpgaconvnet} could improve the utility of our approach.

\section*{Acknowledgments}
The authors would like to thank the NTNU HPC lab and colleagues at Xilinx Research Labs for their support.
This work was supported under the Australian Research Councils Linkage Projects funding scheme (project number LP130101034).

\bibliographystyle{abbrv}
\bibliography{references}

\newcommand{\mytext}{Last accessed:}
\begin{thebibliography}{10}

\bibitem{alemdar2016ternary}
H.~Alemdar, N.~Caldwell, V.~Leroy, A.~Prost-Boucle, and F.~P{\'e}trot.
\newblock {Ternary Neural Networks for Resource-Efficient AI Applications}.
\newblock {\em CoRR}, abs/1609.00222, 2016.

\bibitem{YodaNN}
R.~Andri, L.~Cavigelli, D.~Rossi, and L.~Benini.
\newblock {YodaNN}: An ultra-low power convolutional neural network accelerator
  based on binary weights.
\newblock {\em CoRR}, abs/1606.05487, 2016.

\bibitem{chellapilla2006high}
K.~Chellapilla, S.~Puri, and P.~Simard.
\newblock High performance convolutional neural networks for document
  processing.
\newblock In {\em Proc. ICFHR}. Suvisoft, 2006.

\bibitem{chen2016eyeriss}
Y.-H. Chen, J.~Emer, and V.~Sze.
\newblock Eyeriss: A spatial architecture for energy-efficient dataflow for
  convolutional neural networks.
\newblock In {\em Proc. ACM/IEEE ISCA}. IEEE, 2016.

\bibitem{binarynet}
M.~Courbariaux and Y.~Bengio.
\newblock Binarynet: Training deep neural networks with weights and activations
  constrained to +1 or -1.
\newblock {\em CoRR}, abs/1602.02830, 2016.

\bibitem{esser2016convolutional}
S.~K. Esser, P.~A. Merolla, J.~V. Arthur, A.~S. Cassidy, R.~Appuswamy,
  A.~Andreopoulos, D.~J. Berg, J.~L. McKinstry, T.~Melano, D.~R. Barch, et~al.
\newblock {Convolutional Networks for Fast, Energy-Efficient Neuromorphic
  Computing}.
\newblock {\em CoRR}, abs/1603.08270, 2016.

\bibitem{farabet2009cnp}
C.~Farabet, C.~Poulet, J.~Y. Han, and Y.~LeCun.
\newblock {CNP: An {FPGA}-based processor for convolutional networks}.
\newblock In {\em Proc. IEEE FPL}, pages 32--37. IEEE, 2009.

\bibitem{han2015deep}
S.~Han, H.~Mao, and W.~J. Dally.
\newblock {Deep Compression: Compressing Deep Neural Network with Pruning,
  Trained Quantization and Huffman coding}.
\newblock {\em CoRR}, abs/1510.00149, 2015.

\bibitem{caffepresso}
G.~Hegde, Siddhartha, N.~Ramasamy, and N.~Kapre.
\newblock {CaffePresso: An Optimized Library for Deep Learning on Embedded
  Accelerator-based platforms}.
\newblock In {\em Proc. CASES}, 2016.

\bibitem{iandola2016squeezenet}
F.~N. Iandola, M.~W. Moskewicz, K.~Ashraf, S.~Han, W.~J. Dally, and K.~Keutzer.
\newblock {SqueezeNet: AlexNet-level accuracy with 50x fewer parameters and<
  1MB model size}.
\newblock {\em CoRR}, abs/1602.07630, 2016.

\bibitem{batchnorm}
S.~Ioffe and C.~Szegedy.
\newblock Batch normalization: Accelerating deep network training by reducing
  internal covariate shift.
\newblock In {\em Proc. ICML}, pages 448--456, 2015.

\bibitem{bitwiseneuralnet}
M.~Kim and P.~Smaragdis.
\newblock Bitwise neural networks.
\newblock {\em CoRR}, abs/1601.06071, 2016.

\bibitem{cifar10}
A.~Krizhevsky and G.~Hinton.
\newblock Learning multiple layers of features from tiny images.
\newblock {\em Technical Report}, 2009.

\bibitem{krizhevsky2012imagenet}
A.~Krizhevsky, I.~Sutskever, and G.~E. Hinton.
\newblock Imagenet classification with deep convolutional neural networks.
\newblock In {\em Proc. NIPS}, pages 1097--1105, 2012.

\bibitem{mnist}
Y.~LeCun, L.~Bottou, Y.~Bengio, and P.~Haffner.
\newblock Gradient-based learning applied to document recognition.
\newblock {\em Proc. of the IEEE}, 86(11):2278--2324, 1998.

\bibitem{surveyannhw}
J.~Misra and I.~Saha.
\newblock Artificial neural networks in hardware: A survey of two decades of
  progress.
\newblock {\em Neurocomputing}, 74(1--3):239--255, 2010.

\bibitem{fpgaroofline}
S.~Muralidharan, K.~O'Brien, and C.~Lalanne.
\newblock {A Semi-Automated Tool Flow for Roofline Anaylsis of OpenCL Kernels
  on Accelerators}.
\newblock {\em Proc. Workshop on H2RC}, 2015.

\bibitem{svhn}
Y.~Netzer, T.~Wang, A.~Coates, A.~Bissacco, B.~Wu, and A.~Y. Ng.
\newblock Reading digits in natural images with unsupervised feature learning.
\newblock {\em {NIPS Workshop on Deep Learning and Unsupervised Feature
  Learning}}, 2011.

\bibitem{ovtcharov2015accelerating}
K.~Ovtcharov, O.~Ruwase, J.-Y. Kim, J.~Fowers, K.~Strauss, and E.~Chung.
\newblock Accelerating deep convolutional neural networks using specialized
  hardware, February 2015.

\bibitem{park2016fpga}
J.~Park and W.~Sung.
\newblock {FPGA based implementation of deep neural networks using on-chip
  memory only}.
\newblock In {\em Proc. IEEE ICASSP}, pages 1011--1015. IEEE, 2016.

\bibitem{xnornet}
M.~Rastegari, V.~Ordonez, J.~Redmon, and A.~Farhadi.
\newblock {{XNOR-Net}: ImageNet Classification Using Binary Convolutional
  Neural Networks}.
\newblock In {\em ECCV}, 2016.

\bibitem{ilsvrc}
O.~Russakovsky, J.~Deng, H.~Su, J.~Krause, S.~Satheesh, S.~Ma, Z.~Huang,
  A.~Karpathy, A.~Khosla, M.~Bernstein, A.~C. Berg, and L.~Fei-Fei.
\newblock {ImageNet Large Scale Visual Recognition Challenge}.
\newblock {\em IJCV}, 115(3):211--252, 2015.

\bibitem{deeplearningoverview}
J.~Schmidhuber.
\newblock Deep learning in neural networks: An overview.
\newblock {\em Neural Networks}, 61:85--117, 2015.

\bibitem{simonyan2014very}
K.~Simonyan and A.~Zisserman.
\newblock Very deep convolutional networks for large-scale image recognition.
\newblock {\em CoRR}, abs/1409.1556, 2014.

\bibitem{fpganclnclconvnet}
N.~Suda, V.~Chandra, G.~Dasika, A.~Mohanty, Y.~Ma, S.~B.~K. Vrudhula, J.~Seo,
  and Y.~Cao.
\newblock {Throughput-Optimized OpenCL-based {FPGA} Accelerator for Large-Scale
  Convolutional Neural Networks}.
\newblock In {\em Proc. ACM/SIGDA ISFPGA}, pages 16--25, 2016.

\bibitem{ResiliencyUnderQuantization}
W.~Sung, S.~Shin, and K.~Hwang.
\newblock Resiliency of deep neural networks under quantization.
\newblock {\em CoRR}, abs/1511.06488, 2015.

\bibitem{venieris2016fpgaconvnet}
S.~I. Venieris and C.-S. Bouganis.
\newblock {{fpgaConvNet}: A Framework for Mapping Convolutional Neural Networks
  on {FPGAs}}.
\newblock In {\em Proc. IEEE FCCM}, pages 40--47. IEEE, 2016.

\bibitem{planet}
T.~Weyand, I.~Kostrikov, and J.~Philbin.
\newblock Planet - photo geolocation with convolutional neural networks.
\newblock {\em CoRR}, abs/1602.05314, 2016.

\bibitem{roofline}
S.~Williams, A.~Waterman, and D.~A. Patterson.
\newblock Roofline: an insightful visual performance model for multicore
  architectures.
\newblock {\em Commun. {ACM}}, 52(4):65--76, 2009.

\bibitem{zhang2015optimizing}
C.~Zhang, P.~Li, G.~Sun, Y.~Guan, B.~Xiao, and J.~Cong.
\newblock Optimizing {FPGA}-based accelerator design for deep convolutional
  neural networks.
\newblock In {\em Proc. ACM/SIGDA ISFPGA}, pages 161--170. ACM, 2015.

\bibitem{dorefa}
S.~Zhou, Z.~Ni, X.~Zhou, H.~Wen, Y.~Wu, and Y.~Zou.
\newblock {DoReFa-Net}: Training low bitwidth convolutional neural networks
  with low bitwidth gradients.
\newblock {\em CoRR}, abs/1606.06160, 2016.

\end{thebibliography}

\end{document}